\documentclass[10.5pt,onecolumn]{article}

\usepackage[a4paper,total={6in, 8in}]{geometry}

\usepackage{bm,amsmath,amssymb,amsfonts,graphicx,epsfig,amsthm,color}
\usepackage{bbold,dsfont}

\usepackage[hyphens]{url}
\usepackage{hyperref}
\PassOptionsToPackage{bookmarks=false}{hyperref}

\usepackage[depth=-1]{bookmark}

\usepackage{flushend}
\usepackage[numbers,sort&compress]{natbib}

\usepackage{multirow}
\usepackage{array}
\usepackage{booktabs}
\usepackage{epstopdf}
\usepackage{amsfonts,amssymb}
\usepackage{algorithm}
\usepackage{algpseudocode}
\usepackage{amsmath}

\usepackage{booktabs}       
\usepackage{amsfonts}       
\usepackage{microtype}      

\usepackage{times}
\usepackage{graphicx} 
\usepackage{subcaption} 

\usepackage{wrapfig}

\usepackage{accents}
\makeatletter
\def\wid{\check{{\cc@style\underline{\mskip9.5mu}}}}
\def\Wideubar{\underaccent{{\cc@style\underline{\mskip6mu}}}}
\makeatother

\makeatletter
\def\wideubar{\underaccent{{\cc@style\underline{\mskip9.5mu}}}}
\def\Wideubar{\underaccent{{\cc@style\underline{\mskip6mu}}}}
\makeatother

\makeatletter
\def\widebar{\accentset{{\cc@style\underline{\mskip9.5mu}}}}
\def\Widebar{\accentset{{\cc@style\underline{\mskip6mu}}}}
\makeatother

\theoremstyle{remark}

\allowdisplaybreaks

\begin{document}

\title{
\textbf{Boosting Black-Box Adversarial Attacks with Meta Learning}
}

\author{Junjie Fu, Jian Sun, and Gang Wang\thanks{The work was supported in part by the National Key R\&D Program of China under Grant 2021YFB1714800, in part by the National Natural Science Foundation of China under Grants 62173034, 61925303, 62088101, 61720106011, in part by the CAAI-Huawei MindSpore Open Fund, and in part by the Chongqing Natural Science Foundation under Grant 2021ZX4100027.		
J. Fu, J. Sun, and G. Wang are with the State Key Lab of Intelligent Control and Decision of Complex Systems and the School of Automation, Beijing Institute of Technology, Beijing 100081, China, and also with Beijing Institute of Technology Chongqing Innovation Center, Chongqing 401120, China.
E-mail: 3120200901@bit.edu.cn; sunjian@bit.edu.cn; gangwang@bit.edu.cn.}
}

\maketitle

\begin{abstract}
	Deep neural networks (DNNs) have achieved remarkable success in diverse fields. However, it has been demonstrated that DNNs are very vulnerable to adversarial examples even in black-box settings. A large number of black-box attack methods have been proposed to in the literature. However, those methods usually suffer from low success rates and large query counts, which cannot fully satisfy practical purposes. In this paper, we propose a hybrid attack method which trains meta adversarial perturbations (MAPs) on surrogate models and performs black-box attacks by estimating gradients of the models. Our method uses the meta adversarial perturbation as an initialization and subsequently trains any black-box attack method for several epochs. Furthermore, the MAPs enjoy favorable transferability and universality, in the sense that they can be employed to boost performance of other black-box adversarial attack methods. Extensive experiments demonstrate that our method can not only improve the attack success rates, but also reduces the number of queries compared to other methods.

\end{abstract}

\maketitle

\allowdisplaybreaks


\section{Introduction}
\label{sec:intro}

Deep neural networks (DNNs) have achieved significant success in a number of science and engineering fields in recent years, including image classification, object detection, smart cities, protein folding, and human-machine confrontation \cite{Krizhevsky2012, Li2022, Xie2021, Khan2021, Wanggang2019, Li2021, Chen2022, ren2022mctan, ren2021data}. Yet, while DNNs are widely used in diverse scenarios, serious security risks are revealed. Recently, researchers have shown that DNNs are vulnerable to adversarial examples which lead them to completely wrong classification or recognition results. Adversarial examples are crafted to be misclassified by well-trained classifiers but almost imperceptible to humans. It is very dangerous for DNN systems employed in security-critical applications such as self-driving \cite{Xiao2017}, smart grid \cite{Wang2019,Wu2019}, and financial recognition security system \cite{Sharif2016}, which may lead to huge economic losses or even catastrophic accidents. After realizing this phenomenon, many efforts have been devoted to studying how to generate adversarial samples, which are critical for explaining and understanding adversarial attacks, evaluating the performance of DNNs, as well as improving the robustness of deep learning models.

In general, depending on whether the attacker has knowledge of the target deep model, adversarial attacks can be divided into white-box attacks and black-box attacks. In the white-box setting, the attacker has full access to the model, including the model architecture and model parameters. Leveraging these information, Goodfellow et al. early proposed the Fast Gradient Sign Method (FGSM) \cite{Goodfellow2015} and has aroused considerable attention. After that, a number of adversarial attack methods have been proposed, including DeepFool \cite{Dezfooli2016}, BIM \cite{Kurakin2017}, JSMA \cite{Papernot2016}, C\&W \cite{Carlini2017}, PGD \cite{Madry2018}, and AI-FGSM \cite{Xiao2020}, to name a few. While there exist relatively many well-performing white-box attacks, the more challenging black-box attacks have not been sufficiently studied. In the black-box setting, the attacker cannot access the internal knowledge of the target model, and can only craft adversarial examples using the input-output information. The black-box setting is more practical and closer to real-world applications.

Lately, some progress has been made in black-box attacks. The work \cite{Papernot2017} proposed a black-box attack method based on training surrogate models. Chen et al. \cite{Chen2017} performed black-box attacks by designing a zero-order optimization algorithm which estimate the gradient by querying the model's output at certain input. Building on the two methods above, researchers have proposed some black-box attack methods with higher efficiency and success rates, such as NES \cite{Ilyas2018}, MI-FGSM \cite{Dong2018}, AutoZoom \cite{Tu2018}, P-RGF \cite{Cheng2019}, Sign-OPT \cite{Liu2021}, etc. Although these existing methods have achieved improvement in black-box attacks, there are still drawbacks including e.g., low success rate and low query efficiency. A natural question arises: Can we generate a meta adversarial perturbation which can be used to adversarially manipulative different images while being efficiently trainable on small datasets?

Inspired by MAML \cite{Finn2017}, we find that meta learning can use little data to solve problems analogously. The purpose of meta learning is to train a model on multiple learning tasks, such that it can be quickly adapted to new learning tasks using only a small amount of training data. And MAML propose a model-agnostic algorithm for meta learning to train the model's parameters such that it can achieve rapid adaptation. It is exactly what is needed in the black-box setting. Yuan et al. \cite{Yuan2021} trained a meta adversarial perturbation using MAML in the while-box setting. However, in a black-box model, an attacker cannot obtain the gradient of the model like a white-box model. Therefore it is difficult to simply generalize Yuan's idea to the black-box model since the gradient cannot be obtained directly to train the perturbations by using MAML. New methods still need to be come up with.

Therefore, to improve the query efficiency and boost black-box attacks, in this paper, a new method is proposed for training meta adversarial perturbations and performing black-box attacks. In a nutshell, the main contributions of this paper are summarized as follows.

\begin{itemize}
	\item A new algorithm is proposed to generate meta adversarial perturbations in black-box setting, such that it can use less queries to achieve a higher success rate.
	
	\item A momentum term is added to the zero order optimization process to accelerate attack and reduce the number of queries.
	
	\item The meta adversarial perturbations are shown to have great adaptability and transferability, because the meta adversarial perturbations are trained by the data that is semantically unrelated to the attacked images, the perturbations can still achieve excellent performance.
\end{itemize}

\section{Related Work}

\label{sec:related}
There is a great deal of work on adversarial examples. Here the most relevant works about white-box attacks and black-box attacks are reviewed.

\subsection{White-box attacks}
In white-box attacks, the model is completely known to the attacker, so the attacker can craft the adversarial examples by using internal information of the model directly. Szegedy et al. \cite{Szegedy2013} first proposed a method for generating adversarial examples by solving the optimization problem with L-BFGS.
Soon after, Goodfellow et al. \cite{Goodfellow2015} proposed a fast algorithm requiring only one step of gradient update and achieved good results. After that, a number of adversarial attack methods have been proposed, including DeepFool \cite{Dezfooli2016}, BIM \cite{Kurakin2017}, JSMA \cite{Papernot2016}, C\&W \cite{Carlini2017}, PGD \cite{Madry2018}, and AI-FGSM \cite{Xiao2020}, etc. Recently, Yuan et al. \cite{Yuan2021} proposed a method for training universal adversarial perturbations in a white-box model based on meta learning, and achieve better results than the FGSM method. White-box attacks have already achieved high success rates and relatively high attack efficiencies. However, the black-box setting is more practical and closer to real-world applications than the white-box setting.

\subsection{Black-box attacks}
In the black-box setting, the attacker cannot access the internal knowledge of the target model, and can only craft adversarial examples using the input-output information. Due to the lack of knowledge of the model, black-box attacks are much more difficult than white-box attacks. There exist three main ideas in black-box attacks, which are transfer-based methods, query-based methods and hybrid methods. The three methods mentioned above are introduced.

\begin{itemize}
	\item\textbf{Transfer-based methods}
\end{itemize}

Transfer-based methods perform black-box attacks based on the transferability of the adversarial examples. Transferability means that if an adversarial example can attack the current model, then it is possible to successfully attack other models as well. Papernot et al.  \cite{Papernot2017} trained a surrogate model to perform black-box attacks with a synthetic dataset. When performing an attack, the attacker first generates an adversarial example on the surrogate model by using a white-box method, and then exploit the transferability of the adversarial example to perform the black-box attack on the target model. Later on, Dong et al. \cite{Dong2018} proposed MI-FGSM by adding a momentum term to the iterative process, which had higher attack efficiency. 

Due to the absence of estimated black-box model information, transfer-based methods have low success rates and large distortions. But they have high efficiency since they don't require lots of queries on the black-box model.

\begin{itemize}
	\item\textbf{Query-based methods}
\end{itemize}

Query-based methods perform black-box attacks based on estimating gradient of models by continuously querying the models. Chen et al. \cite{Chen2017} proposed an zero-order optimization algorithm named ZOO to estimate gradient of models by performing finite difference on the query results. And then, Ilyas et al. proposed NES \cite{Ilyas2018} to perform attack in a query limit condition, which improved query efficiency a lot and achieved higher success rate with lower queries. Tu et al. \cite{Tu2018} used encoding-decoding method to reduce the query dimension of input and performed zero-order optimization with RGF \cite{Nesterov2017} method. 

Query-based methods require a large number of queries on the model, so they're very inefficient. But they have much higher success rates and lower distortions than transfer-based methods, since they access the gradient by a direct query to the target model.

\begin{itemize}
	\item\textbf{Hybrid methods}
\end{itemize}

Building on the two methods above, hybrid methods integrate the advantages of the two methods, which use surrogate model to acquire transfer-based prior knowledge and zero-order optimization to estimate information of target model more accurately. So hybrid methods also have higher success rates and fewer queries. Dong et al. \cite{Cheng2019} proposed a hybrid method named P-RGF, which used the gradient of surrogate model as prior knowledge to guide the query direction of RGF and obtained the same success rate as RGF with fewer queries.

\subsection{Non-targeted and targeted attacks}
According to the purpose of attack, adversarial attacks can be divided into non-targeted attacks and targeted attacks. Non-targeted attacks require that the model's prediction of input image is different from its ground truth. While targeted attacks not only need to meet the above requirements, but also require the image to be classified into the specified target class. Obviously, targeted attacks are more difficult and more threatening. In the existing research on black-box attacks, the non-targeted attacks can be easily performed with high success rate. But the target attacks are still unsatisfactory, which still need large queries and have low success rates. Therefore, this paper focuses on targeted attacks mainly.

\section{Methodology} 
\label{sec:formulation}

Our method belongs to hybrid methods. Unlike other hybrid methods that use surrogate model gradients as priori knowledge, our method aims at training a more universal adversarial perturbation as an initial value by using surrogate models based on the transferability and adaptation of adversarial examples. According to the setting of black-box attacks, we want to generate a meta adversarial perturbation which can be used to adversarially manipulative different images while being efficiently trainable on small datasets. Inspired by the excellent performance of MAML \cite{Finn2017} in many tasks, meta learning method is used and the black-box attacks are modeled analogously. Therefore, due to the transferability of adversarial examples, surrogate models are used to generate meta adversarial perturbations (MAPs) based on meta learning methods.

In this section, a method is first proposed which uses meta learning to train meta adversarial perturbations on surrogate models. Then a black-box attack method is proposed which uses the zero-order optimization algorithm to estimate the gradient of target models and adds a momentum term to accelerate attack.

\subsection{MAPs training}
Formally, consider the black-box model as $f_\theta$, and the surrogate models as $f_i,\ (i=1,2,...,n)$ with known architectures and parameters. A meta adversarial perturbation is denoted as $v$ which is randomly initialized. $L$ denotes a loss function which can be a cross-entropy loss, a margin loss or some other loss, and the cross-entropy loss is used in this paper for convience. $\mathbb{D}$ denotes the train set which is used to generate MAPs. When adapting to a new batch of data points $\mathbb{B}=\left\lbrace x^{i},y^{i}\right\rbrace \sim\mathbb{D}$, the MAP $v$ becomes $v'$ which is adaptable and kind of adversarial. We want to find a universal meta adversarial perturbation $v$ that can be quickly adapted to any new data point to become a good initial value for attack. In targeted attacks, a perturbation $v$ is supposed to satisfy
\begin{equation}
	\label{eq1}
	f\left(x+v\right) = t, \ \   \text{for most} \ x\sim \mu
\end{equation}
Where $t$ is the target label and $\mu$ is the data distribution. 

It can be found that MAPs are image-agnostic as a trained MAP can adapt to all the data points, no matter whether this point is used as training data for the MAP or not. This is a very good property for a black-box attack setting since the trained MAP can be used directly without contacting the black-box model, which significantly reduces the queries to the black-box model and improves the attack efficiency. In the following, our method is presented for training MAPs in the black-box setting.

The MAPs are trained by using a gradient-based iterative method. So, it is difficult to train MAPs directly on the victim model since the model is black-box and its true gradient cannot be obtained. Exploiting the transferability of adversarial examples and the model-agnosticism of meta learning, we utilize meta learning to train MAPs on the surrogate models and we believe that the MAPs are still well adapted and adversarial in the black-box model. In our method, multi-step gradient descent is used to update perturbation on new data points with surrogate models $f_i,\ (i=1,2,...,n)$
\begin{equation}
	\label{eq2}
	v'\gets v-\alpha\cdot\frac{1}{n}\sum_{i=1}^n\nabla_{v}L\left(f_{i},\mathbb{B}+v,t\right)
\end{equation}
Where the learning rate $\alpha$ is a hyperparameter and usually set to a small value. The perturbation $v'$ is called adapted perturbation since it has adapted the data points in mini-batch $\mathbb{B}$ and already somewhat adversarial.

The meta perturbation is updated by minimizing the loss corresponding to $v$ in the new mini-batch of data $\mathbb{B'}$ by adding the adapted perturbation $v'$. The optimization objective can be described as

\begin{equation}
	\label{eq3}
	\min_{v}\sum_{\mathbb{B'}\sim\mathbb{D}}L(f_\theta, \mathbb{B'}+v',t)
\end{equation}

The optimization objective as formula~(\ref{eq3}) is performed over the perturbation $v$, while the objective is computed on the adapted perturbation $v'$. In fact, our aim is to first find a better adapted perturbation $v'$ in data $\mathbb{B}$ and then optimize the meta adversarial perturbation $v$ at the new data point iteratively. Then, a maximally effective adversarial perturbation $v$ can be obtained which can perform attack with a high success rate and acquire better adaptability.

\begin{algorithm}[!htb]  
	\caption{MAP training}  
	\label{alg1}  
	\begin{algorithmic}[1]  
		\Require  
		Trainset $\mathbb{D}$; learn rate $\alpha$; learn rate $\beta$; surrogate model $f_i$; loss function $L$; project $\Pi_{\epsilon}$; target label $t$.
		\Ensure  
		Meta adversarial perturbation $v$.
		\State Randomly initialize $v$;
		\While{not done}
		\For{minibatch $\mathbb{B}=\left\lbrace x^{i},y^{i}\right\rbrace \sim\mathbb{D}$} 
		\For{all $f_i$}
		\State Evaluate $\nabla_{v}L\left(f_{i},\mathbb{B}+v,t\right)$ using minibatch $\mathbb{B}$ \Statex \qquad \qquad \,  with perturbation $v$;
		\State Compute adapted perturbation with gradient \Statex \qquad \qquad \,\, descent:
		\State $v'\gets v-\alpha\cdot\frac{1}{n}\sum_{i=1}^n\nabla_{v}L\left(f_{i},\mathbb{B}+v,t\right)$;
		\EndFor
		\State sample a minibatch of data $\mathbb{B'}$ from $\mathbb{D}$;
		\For{all $f_i$}
		\State Evaluate $\nabla_{v'}L\left(f_{i},\mathbb{B'}+v',t\right)$ using minibatch $\mathbb{B}$ \Statex \qquad \qquad \,  with adapted perturbation $v'$;
		\State update: $v\gets v-\beta\cdot\frac{1}{n}\sum_{i=1}^n\nabla_{v'}L\left(f_{i},\mathbb{B'}+v',t\right)$;
		\EndFor
		\State project $v\gets\Pi_{\epsilon}\left(v\right) $;
		\EndFor
		\EndWhile \\
		\Return $v$.
	\end{algorithmic}  
\end{algorithm}

Stochastic gradient descent is utilized to update the meta adversarial perturbation $v$
\begin{equation}
	\label{eq4}
	v\gets v-\beta\cdot\frac{1}{n}\sum_{i=1}^n\nabla_{v'}L\left(f_{i},\mathbb{B'}+v',t\right)
\end{equation}
Where $\beta$ is meta learning rate, which is a hyperparameter like $\alpha$. The detailed procedure of MAP training is presented in Algorithm~\ref{alg1}. At line 14, a projection operation is performed on the updated perturbations to restrict them in a norm bound controlled by $\epsilon$. A smaller $\epsilon$ makes an attack less visible to humans.

\subsection{Query-efficient attack}
In this section, a query-efficient black-box attack algorithm is designed using MAPs trained in section 3.1. When performing an attack, the trained MAP is first added to the input image. Due to the strong effectiveness of MAP, some images can successfully attack the black-box model directly only after adding MAP without other operations. If the attack is not successful, the attack is continuously performed using zero-order optimization algorithms to estimate the gradient of the black-box model. Here the gradient of the model is estimated by using RGF method \cite{Nesterov2017} as

\begin{equation}
	\label{eq5}
	\hat{g}=\frac{1}{q}\sum_{k=1}^q\frac{f(x+\sigma u_k,y)-f(x,y)}{\sigma}\cdot u_k
\end{equation}

Then the input image $x$ is updated iteratively using the estimated gradient $\hat{g}$ under $L_{\infty}$ norm. To accelerate the efficiency of the attack, a momentum term is added which is the difference between two adjacent iterations of the image to the iterative process. This momentum acceleration method was first proposed by Chen et al. \cite{Chen2021} and was proved to be effective. The update of image $x$ is as

\begin{equation}
	\label{eq6}
	x_{i+1}=x_i-\gamma\cdot sign(\hat{g})+\eta \cdot (x_i-x_{i-1})
\end{equation}
where $\gamma$ is step size and $\eta$ is momentum coefficient, they both should be taken to a reasonably small value. $sign(\cdot)$ denotes a symbolic function.

After each update, a projection operation is performed on the current image $x$ as follows

\begin{equation}
	\label{eq7}
	x=\Pi_{\varepsilon}(x)
\end{equation}
where $\Pi_\varepsilon(x)$ projects vector x onto the norm ball $\|x\|_2\le \varepsilon$. Algorithm~\ref{alg2} outlines the key steps of our proposed black-box meta adversarial attack. 

\begin{algorithm}[!htb]  
	\caption{Black-box meta adversarial attack}  
	\label{alg2}  
	\begin{algorithmic}[1]  
		\Require  
		Trained MAP $v_t$ for target label $t$; black-box model $f_\theta$; input image $x$; smooth coefficient $\sigma$; step size $\gamma$; momentum coefficient $\eta$; iterations $I$; number of queries $q$; project $\Pi_{\varepsilon}$.
		\Ensure  
		Adversarial example $x_{adv}$.
		\State $x\gets x+v$;
		\If{$f_\theta(x)=t$}
		\State $x_{adv}=x$;
		\Else
		\For{$i=1$ to $I$}
		\State $\hat{g}=\frac{1}{q}\sum_{k=1}^q\frac{f(x+\sigma u_k,y)-f(x,y)}{\sigma}\cdot u_k$;
		\State $x'=x_i-\gamma\cdot sign(\hat{g})+\eta \cdot (x_i-x_{i-1})$;
		\State $x=\Pi_{\varepsilon}(x')$;
		\If{$f_\theta(x)=t$}
		\State $x_{adv}=x$;
		\State break;
		\Else
		\State$x_{i+1}=x$;
		
		\EndIf
		
		\EndFor
		\EndIf \\
		\Return $x_{adv}$.
	\end{algorithmic}  
\end{algorithm}

\section{Experiments}
\label{sec:results}

In this section, we present the experimental setup and results, and we demonstrate the effectiveness of our method by comparing it with some state-of-the-art black-box attack methods. 

\subsection{Experimental setup}
The representative benchmark dataset CIFAR10 was used to assess our method. The CIFAR10 consists of RGB images with 10 classes containing animal and transportation. The size of the image is $32\times32\times3$. It includes 50,000 images from the train set and 10,000 images from the test set. The surrogate models and victim models used in our experiment were all trained on the train set, but had different architectures and parameters.
Five trained models were used in our experiment, including VGG13, VGG16,  GoogleNet, ResNet18 and ResNet34, and all of these models exceeded 85\% accuracy on the test set. When performing attack, one model was chosen as the black-box model and others were used to train MAPs as white-box models. The parameters were set as follows: learn rate $\alpha=\beta=1.5\times10^{-5}$ and iterated 20 epochs in MAP training, while smooth coefficient $\sigma=0.1$, number of queries $q=14$, step size $\gamma=0.015$ under $L_{\infty}$ norm and momentum coefficient $\eta=0.01$ in black-box attack.

Since the difficulties of non-targeted attacks in the black-box setting have been well addressed, our experiments mainly focused on targeted attacks. For non-targeted attacks, our method can also be applied with a slight modification.

\subsection{Targeted attacks}
100 images were selected from each of the 10 classes, i.e. 1000 images to form the dataset $\mathbb{D}$, and the other 900 images were selected from 9 classes except the target class as the attacked dataset. When training the MAPs, the minibatches were set to $\mathbb{B}=600$ and $\mathbb{B'}=1000$. Then stochastic gradient descent was used to update MAPs for the target attack. Our method was compared with RGF \cite{Nesterov2017},  NES \cite{Ilyas2018} and P-RGF \cite{Cheng2019} with the same attack settings to be fair. Since our method added a MAP to the input image, a random perturbation was also added with the same distortion as a comparison. We chose target label $t=6$ and set 600 as the maximal number of queries on the black-box model. The results of the attack are as Table~\ref{tab1}. 

\begin{table}[!htb]
	\renewcommand\arraystretch{1.2}
	\centering
	\caption{Targeted attacks}
	\label{tab1}
	\begin{tabular}{c|c|c|c|c}
		\hline
		Model & Method & Success & Queries & Avg. $L_2$ norm\\ 
		\hline
		\multirow{5}*{VGG13}&RGF&85.11\%&270.53&3.32 \\ 
		\cline{2-5} &Random &88.44\%&240.43&4.00\\
		\cline{2-5} &P-RGF &82.00\%&137.16&\textbf{3.18}\\
		\cline{2-5} &NES &76.56\%&310.48&3.62\\
		\cline{2-5} &Ours &\textbf{96.33\%}&\textbf{130.45}&3.23\\
		\hline
		\multirow{5}*{VGG16}&RGF&87.11\%&272.03&3.32 \\ 
		\cline{2-5} &Random &89.33\%&249.51&4.03\\
		\cline{2-5} &P-RGF &87.22\%&120.50&2.96\\
		\cline{2-5} &NES &78.33\%&305.93&3.58\\
		\cline{2-5} &Ours &\textbf{97.11\%}&\textbf{108.09}&\textbf{2.80}\\
		\hline
		\multirow{5}*{GoogleNet}&RGF&97.11\%&243.07&\textbf{3.17} \\ 
		\cline{2-5} &Random &98.22\%&221.36&3.92\\
		\cline{2-5} &P-RGF &98.89\%&\textbf{129.44}&3.53\\
		\cline{2-5} &NES &88.33\%&291.73&3.53\\
		\cline{2-5} &Ours &\textbf{99.00\%}&148.69&3.52\\
		\hline
		\multirow{5}*{ResNet18}&RGF&98.33\%&202.98&2.90 \\ 
		\cline{2-5} &Random &98.67\%&184.58&3.63\\
		\cline{2-5} &P-RGF &99.00\%&107.18&2.96\\
		\cline{2-5} &NES &95.00\%&238.49&3.18\\
		\cline{2-5} &Ours &\textbf{99.67\%}&\textbf{80.67}&\textbf{2.75}\\
		\hline
		\multirow{5}*{ResNet34}&RGF&94.00\%&233.56&\textbf{3.09} \\ 
		\cline{2-5} &Random &95.67\%&197.14&3.80\\
		\cline{2-5} &P-RGF &96.89\%&134.11&3.29\\
		\cline{2-5} &NES &86.89\%&268.90&3.37\\
		\cline{2-5} &Ours &\textbf{98.33\%}&\textbf{108.42}&3.45\\
		\hline
	\end{tabular}
\end{table}

Table~\ref{tab1} shows the success rate of black-box attacks, the average number of queries needed to generate an adversarial example and the average distortion of all adversarial examples. It can be seen that our method generally leads to higher success rates, fewer queries and relatively smaller distortions than other methods. When the number of queries is limited, our method has around 10\% improvement in success rates over other methods in some models. And our method usually uses fewer queries to achieve the same success rate. Although the distortions caused by our method are slightly larger in some models, they are still at a very small level.
The results also show that the MAPs generated by our method have great transferability and adaptation on the new data points, since they can
work well on the attacked dataset and perform the black-box attack well. Experiments prove that our method is much more queries-efficeint and more effective than other methods. 

\begin{figure}[!htb]
	\centering
	\includegraphics[width=8cm]{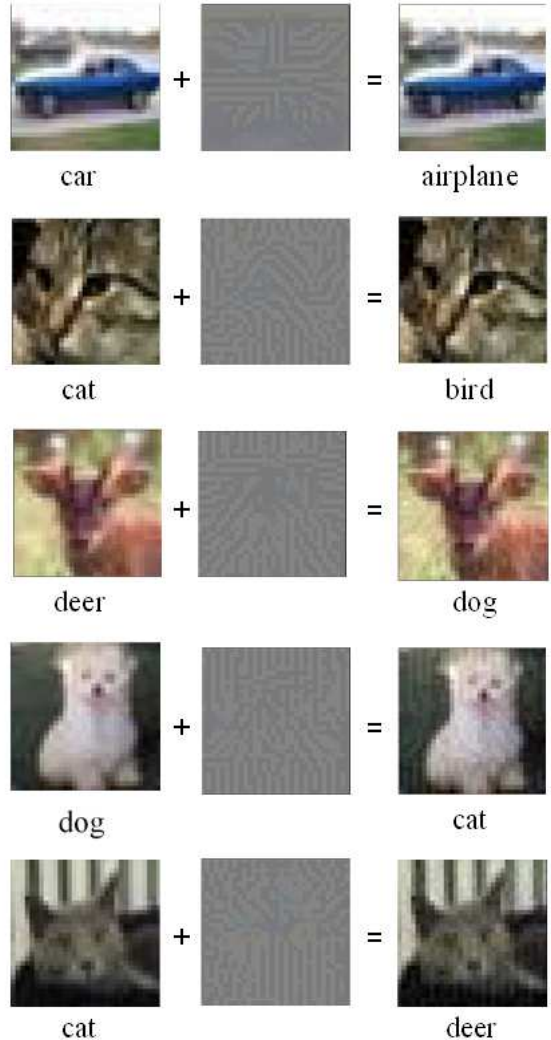}
	\caption{Adversarial examples generated by our method}
	\label{fig1}
\end{figure}

Fig.~\ref{fig1} illustrates the adversarial examples generated by our method. The left column shows the original images to be attacked, the middle column shows the adversarial perturbations generated by our method and the right column shows the adversarial examples corresponding to the target class. It can be seen that the distortions caused by our method are almost invisible.

\subsection{Adaptability and universality}

In this section, the transferability and universality of MAPs has been further investigated. It is intriguing that whether the performance of MAPs in the attack is sensitive to the selection of training data. The animal images in CIFAR10 are chosen as the training dataset while the transportation images as the attacked dataset. This is a great test for the adaptability and versatility of MAPs, because the MAPs have never learnt the features of the attacked images during the training process. If MAPs can still obtain great attack results under such unfavourable conditions, it can be demonstrated that the MAPs trained by our method are well adapted and universal. Here 100 images were selected from the 6 classes of animals to form the dataset $\mathbb{D}$, and the other 300 images were selected from the classes of transportation except the target class. When training the MAPs, the minibatches were set to $\mathbb{B}=360$ and $\mathbb{B'}=600$. Our method was compared with the methods in section 4.1. We chose target label $t=9$ and set 600 as the maximal number of queries on the black-box model. The results of the attack are as Table~\ref{tab2}.

\begin{table}[!htb]
	\renewcommand\arraystretch{1.2}
	\centering
	\caption{Adaptability and universality}
	\label{tab2}
	\begin{tabular}{c|c|c|c|c}
		\hline
		Model & Method & Success & Queries & Avg. $L_2$ norm\\ 
		\hline
		\multirow{5}*{VGG13}&RGF&86.33\%&244.16&3.14 \\ 
		\cline{2-5} &Random &85.33\%&230.38&4.86\\
		\cline{2-5} &P-RGF &83.00\%&\textbf{111.83}&\textbf{2.90}\\
		\cline{2-5} &NES &75.67\%&286.17&3.46\\
		\cline{2-5} &Ours &\textbf{96.33\%}&158.67&3.39\\
		\hline
		\multirow{5}*{VGG16}&RGF&80.33\%&243.37&3.13 \\ 
		\cline{2-5} &Random &79.00\%&227.02&4.95\\
		\cline{2-5} &P-RGF &79.33\%&\textbf{115.87}&\textbf{2.76}\\
		\cline{2-5} &NES &69.33\%&281.14&3.48\\
		\cline{2-5} &Ours &\textbf{95.67\%}&165.81&3.46\\
		\hline
		\multirow{5}*{GoogleNet}&RGF&97.67\%&265.29&\textbf{3.30} \\ 
		\cline{2-5} &Random &98.33\%&252.49&4.98\\
		\cline{2-5} &P-RGF &96.33\%&\textbf{177.86}&3.43\\
		\cline{2-5} &NES &82.67\%&316.18&3.77\\
		\cline{2-5} &Ours &\textbf{98.67\%}&186.98&3.72\\
		\hline
		\multirow{5}*{ResNet18}&RGF&93.33\%&280.94&\textbf{3.41} \\ 
		\cline{2-5} &Random &95.33\%&265.21&4.95\\
		\cline{2-5} &P-RGF &91.67\%&242.34&3.51\\
		\cline{2-5} &NES &77.67\%&329.93&3.75\\
		\cline{2-5} &Ours &\textbf{96.33\%}&\textbf{230.54}&4.01\\
		\hline
		\multirow{5}*{ResNet34}&RGF&91.33\%&285.94&\textbf{3.43} \\ 
		\cline{2-5} &Random &90.67\%&269.19&5.00\\
		\cline{2-5} &P-RGF &91.00\%&218.18&3.58\\
		\cline{2-5} &NES &77.00\%&339.51&3.81\\
		\cline{2-5} &Ours &\textbf{95.00\%}&\textbf{201.25}&3.76\\
		\hline
	\end{tabular}
\end{table}

As shown in Table~\ref{tab2}, our method achieves a higher success rate than other methods without knowing the features of the attacked class. Our method achieves the highest attack success rate in all models, and the success rate of our method is even improved by more than 10\% in some models. And our method still maintains few queries for the black-box model. The queries of our methods are kept to the fewest or nearly the fewest. Due to the absence of feature knowledge, our method causes slightly larger distortions in some models. However, our method still outperforms other methods in terms of overall performance. It also demonstrates the excellent adaptability and universality of the MAPs generated by our method. 

From the experimental results, it can be concluded that the MAPs generated by our method have excellent adaptability and universality in different network structures and different data points. They can execute attacks independently of the network and data, which is very suitable for the black-box setting. Therefore, it can be said that the MAPs generated by our method are truly universal adversarial perturbations for black-box attacks.

\section{Conclusions}
\label{sec:conc}

In this paper, we have proposed a new black-box attack method to utilize the meta adversarial perturbations (MAPs) for boosting black-box attack. The method contains two stages: (1) training MAPs on the surrogate models using meta learning method; (2) performing black-box attack with the trained MAPs and estimating the gradient of the black-box models efficiently. The experimental results show that our method can achieve higher success rates and fewer queries than other methods. And the MAPs trained by our method have excellent transferability and universality. In future work, we will try to train more efficient MAPs on large datasets and further boost black-box attacks.

\bibliographystyle{IEEEtranS}
\bibliography{MAP}

\end{document}